\documentclass{ieeeaccess}
\usepackage{cite}
\usepackage{amsmath,amssymb,amsfonts}
\usepackage{algorithmic}
\usepackage{graphicx}
\usepackage{textcomp}
\usepackage{booktabs}
\usepackage{nicefrac}
\usepackage{siunitx}
\usepackage{tikz}
\usepackage[implicit=false, hidelinks]{hyperref}

  \NewSpotColorSpace{PANTONE}
  \AddSpotColor{PANTONE} {PANTONE3015C} {PANTONE\SpotSpace 3015\SpotSpace C} {1 0.3 0 0.2}
  \SetPageColorSpace{PANTONE}%

\def\BibTeX{{\rm B\kern-.05em{\sc i\kern-.025em b}\kern-.08em
    T\kern-.1667em\lower.7ex\hbox{E}\kern-.125emX}}
\begin{document}
\doi{10.1109/ACCESS.2021.3116304}

\title{Point Transformer}
\author{\uppercase{Nico Engel}\authorrefmark{1},
\uppercase{Vasileios Belagiannis\authorrefmark{1} and Klaus Dietmayer\authorrefmark{1}}}
\address[1]{Institute of Measurement, Control and Microtechnology, Ulm University, Albert-Einstein-Allee 41, 89081 Ulm, Germany (e-mail: firstname.lastname@uni-ulm.de)}

\markboth
{N. Engel \headeretal: Point Transformer}
{N. Engel \headeretal: Point Transformer}

\corresp{Corresponding author: Nico Engel (e-mail: nico.engel@uni-ulm.de).}

\definecolor{color_pcl}{rgb}{0.784,0.784,0.784}
\definecolor{color_selection}{rgb}{0.0784,0.0784,0.0784}

\newcommand\copyrighttext{%
	\scriptsize \copyright~2021 IEEE. Personal use of this material is permitted. Permission from IEEE must be obtained for all other uses, in any current or future media, including reprinting/republishing this material for advertising or promotional purposes, creating new collective works, for resale or redistribution to servers or lists, or reuse of any copyrighted component of this work in other works.}%
\newcommand\copyrightnotice{%
\begin{tikzpicture}[remember picture,overlay]
\node[anchor=south,yshift=18pt,xshift=0.25cm,scale=0.8] at (current page.south) {{\parbox{\dimexpr\textwidth-\fboxsep-\fboxrule\relax}{\copyrighttext}}};
\end{tikzpicture}%
}

%
%
\begin{abstract}
In this work, we present Point Transformer, a deep neural network that operates directly on unordered and unstructured point sets. We design Point Transformer to extract local and global features and relate both representations by introducing the local-global attention mechanism, which aims to capture spatial point relations and shape information. For that purpose, we propose SortNet, as part of the Point Transformer, which induces input permutation invariance by selecting points based on a learned score. The output of Point Transformer is a sorted and permutation invariant feature list that can directly be incorporated into common computer vision applications. We evaluate our approach on standard classification and part segmentation benchmarks to demonstrate competitive results compared to the prior work. Code is publicly available at: \url{https://github.com/engelnico/point-transformer}
\end{abstract}


\titlepgskip=-15pt

\maketitle
\copyrightnotice

\section{Introduction}\label{seq:01_intro}
\begin{figure*}[t]
\begin{center}
\includegraphics[width=0.9\textwidth]{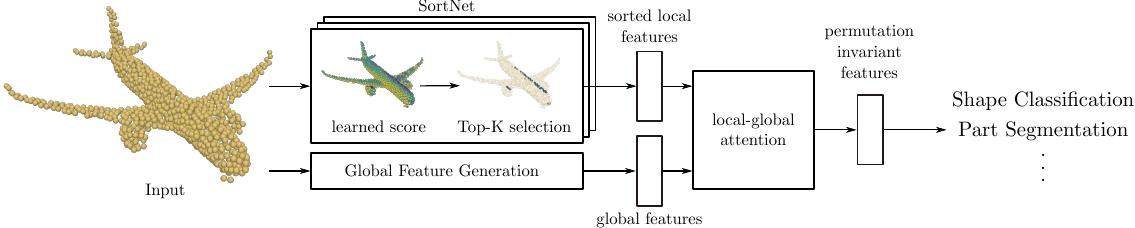}
\caption{Overview of the Point Transformer Pipeline. A point cloud serves as input to our network from which local and global features are extracted. We sort local features using SortNet, a module that focuses on important points based on a learned score. We then employ local-global attention to relate global and local features. We aim to capture geometric relations and shape information. The resulting feature representation is permutation invariant and can be used for common computer vision tasks. }
    \label{fig:overview}
    \end{center}
\end{figure*}

Processing 3D point sets using deep neural networks has become very popular the past few years. The three-dimensional information has a wide range of applications in autonomous driving~\cite{lang2019pointpillars, engel2018deep, engel2019deeplocalization, simon2020stickypillars, horn2020deepclr, wiederer2020} and computer vision~\cite{qi2017pointnet++,qi2019deep}. However, training neural networks on point sets is not trivial. First, point sets are unordered, thus require the neural network to be permutation invariant. Second, the number of points in the set is usually dynamic and unstructured. Finally, the network needs to be robust against rotation and translation to operate in the metric space, and since the points describe objects, the network needs to capture the spatial relations between the points.

Standard neural architectures, such as convolutional neural networks (CNN), have shown promising results for structured data. For that reason, several point set processing approaches attempt to transform the points into regular representations such as voxel grids~\cite{maturana2015voxnet,wu20153d} or rendered views of the point clouds~\cite{qi2016volumetric,su2015multi}. However, transforming the point sets leads to loss of shape information as geometric relations between points are removed. Furthermore, these methods suffer from high computational complexity due to the sparsity of the 3D points. 
To address these limitations, there is another family of approaches that act directly on the point set. The main idea is to process each point individually with a multi-layer perceptron (MLP) and then fuse the representation to a vector of fixed size with a set pooling operation over a latent feature space~\cite{qi2017pointnet,qi2017pointnet++}.
Set pooling is a symmetric function that is permutation invariant. Additionally, under certain conditions, set pooling acts as a universal set function approximator~\cite{zaheer2017deep}. 
Nevertheless, Wagstaff \textit{et al}.~\cite{wagstaff2019limitations} argue that reducing the latent representation to a vector of fixed length can be impractical since the cardinality of the input set is usually not considered. Thus, the capacity of the vector may not be sufficient enough to capture the spatial relations of the point set which may reduce the overall performance. Therefore, the set pooling mechanism can become a bottleneck for point processing networks. 

Our goal and motivation stems from removing the set pooling method and overcoming the aforementioned bottleneck, while still achieving a permutation invariant representation that models the point set relations in terms of object shape and geometric dependencies. Therefore, it is necessary to introduce a symmetric set function that replaces traditional set pooling operations. For that, we adapt the attention mechanism~\cite{bahdanau2014neural}, which was originally introduced for natural language processing, that is used to weight and score sequences (words) based on learned importance.
To our understanding, we face a similar problem in 3D point processing, given that we need to relate representations of the input points to capture and describe the object's shape. Additionally, attention itself does not depend on the input ordering, i.e. it is permutation-invariant,  as it is comprised of matrix multiplication and summation only, which makes it well-suited for our problem. However, the output is still unordered, thus, directly processing the output of attention for standard computer vision tasks is not possible. 
Consequently, our goals can be outlined as follows: 
\begin{itemize}
    \item Avoid the bottleneck that can occur while employing set pooling operations~\cite{wagstaff2019limitations}. 
    \item Present a novel permutation invariant network architecture that adapts the popular and prevalent attention mechanism for 3D point processing.
    \item Demonstrate superior performance compared to traditional set pooling methods to justify the use of attention and reinforce the claims made by Wagstaff \textit{et al}.
\end{itemize}

To address these problems, we propose \textbf{SortNet}, a permutation invariant network module, that learns ordered subsets of the input with latent features of local geometric and spatial relations. For that, we learn important key points, which we call top-k selections, that replace the set pooling operation.
Since current state-of-the-art methods have shown that aggregating local and global information increases the network's capabilities of capturing context information~\cite{qi2017pointnet++,liu2019point2sequence,li2018so}, we employ SortNet to generate local features of the point cloud. Moreover, global features of the entire point cloud are related to the sorted local features using \textbf{local-global attention}. Local-global attention attends both feature representations to capture the underlying shape. Since the local features are ordered, the output of local-global attention is ordered and permutation invariant; and thus it can be used for a variety of visual tasks such as shape classification and part segmentation. An overview of our network is outlined in Fig.~\ref{fig:overview}. Since we aim to process 3D point sets using the ideas proposed by the Transformer network architecture~\cite{vaswani2017attention}, we took inspiration from~\cite{lee2019set}, and name our network \mbox{\textbf{Point Transformer}.}

Overall, our contributions can be summarized as follows:
\begin{itemize}
    \item We propose \textbf{Point Transformer}, a neural network that uses the multi-head attention mechanism and operates directly on unordered and unstructured point sets.
    \item We present \textbf{SortNet}, a key component of Point Transformer, that induces permutation invariance by selecting points based on a learned score.
    \item We evaluate Point Transformer on two standard benchmarks and show that it delivers competitive results.
\end{itemize}

\section{Related Work}\label{seq:02_rel_work}
Below, we discuss approaches that process 3D points and are related to our work.

\subsection{Point set processing}

Point clouds are irregular and unordered sets of points with a variable amount of elements, thus applying standard neural networks on 3D points is not possible. For that reason, previous approaches rely on transforming the point sets into an ordered representation, such as voxel grids. The metric space is discretized into small regions (voxels), which are labeled as occupied if a point lies inside the voxel. Then, 3D convolutional networks (CNN) can be easily applied to the voxel-based representation~\cite{maturana2015voxnet,wu20153d,zhou2018voxelnet}. This pre-processing, however, reduces the resolution as multiple points are combined into a single voxel and thus damages important spatial relations of the metric space. Furthermore, voxelization increases the memory requirements and computational complexity due to the sparsity of the 3D points. To address these limitations, multiple extensions have been proposed that try to leverage the sparsity of 3D data~\cite{wang2015voting,li2016fpnn,riegler2017octnet}, but still fail to process large amounts of input points.

\textbf{View-based methods:} In contrast to building voxel grids, a lot of research has been conducted on rendering point clouds into 2D images, i.e.~structured representation of the underlying 3D shape. Then, working with traditional CNNs is possible~\cite{shi2015deeppano,su2015multi}. Since shape information can be occluded by rendering point clouds from a specific viewpoint, multi-view approaches have been proposed that render multiple images from different angles~\cite{su2015multi,kanezaki2018rotationnet,su2018deeper,qi2016volumetric}. Even though images are rendered from different views, the model still fails to capture all geometric and spatial relations. To this day, multi-view approaches achieve impressive results on standard 3D benchmarks. However, the transformation from sparse 3D points into images increases computational complexity as well as required memory. 

\textbf{Shape-based methods:} PointNet~\cite{qi2017pointnet} is a pioneering network architecture that operates directly on 3D point sets, and it is invariant to input point permutations. Therefore, a transformation into a structured representation is no longer necessary.
PointNet uses a multi-layer perceptron (MLP) with shared weights that encodes spatial features to each input point separately. Then, a symmetric function, e.g.~max pooling, is applied to the latent features to induce permutation invariance and create a global feature representation of the input. PointNet established the de facto standard for point processing that many state-of-the-art approaches still rely on~\cite{lang2019pointpillars,qi2018frustum}. However, it is not able to encode and capture local information, since the max pooling operation induces permutation invariance, but also destroys local structures and relations of the points in metric space. To address this issue, Qi \textit{et al}.~proposed the improved PointNet++~\cite{qi2017pointnet++} architecture, a hierarchical model that abstracts the input points with every layer to produce sets with fewer elements. First, centroids of local regions are sampled using hand-crafted algorithms, then local features are encoded to the centroids by exploring the local neighborhood. Thus, allowing the network to capture fine-grained patterns and improving the performance on current datasets.
A general approach related to unordered sets was introduced by Zaheer \textit{et al}.~\cite{zaheer2017deep} demonstrating the capabilities of pooling operations to induce permutation invariance. Importantly, they prove that the set pooling method is a universal approximator for any set function. In general, problems arise with set pooling when the reduced feature vector lacks the capacity to capture important geometric relations. Our work addresses this limitation with a network topology that encodes the entire point cloud by relating local information with the global shape structure.

\textbf{Convolutions on Point Clouds:} Classic convolutional neural networks require the input data to be ordered, such as images or voxel grids. Since points are unstructured, an active research area is the definition of convolution operations that can operate on irregular 3D point sets such as KPConv~\cite{thomas2019kpconv}, SpiderCNN~\cite{xu2018spidercnn} or PointCNN~\cite{li2018pointcnn}. These methods achieve state-of-the-art performance on a variety of tasks. However, due to the irregularities of the shape and point density, point convolutions are usually hard to design and the kernel needs to be adapted for different input data~\cite{guo2020deep}.

\subsection{Attention}
Attention itself has its origin in natural language processing~\cite{bahdanau2014neural,luong2015effective}. Traditionally, encoder-decoder recurrent neural networks (RNN) were used for machine translation applications, where the last hidden state is used as the context vector for the decoder to sequentially produce the output. The problem is that dependencies between distant inputs are difficult to model using sequential processing. Bahdanau \textit{et al}.~\cite{bahdanau2014neural} introduced the attention mechanism that takes the whole input sequence into account by taking the weighted sum of all hidden states and additionally, models the relative importance between words. Vaswani \textit{et al}.~\cite{vaswani2017attention} improved the attention mechanism by introducing multi-head attention and proposing an encoder-decoder structure that solely relies on attention instead of RNNs or convolutions. Therefore, they reduce the computational complexity. In this work, multi-head attention is the basis for Point Transformer.

\textbf{Attention with point cloud processing:} Neural networks that rely on attention achieved impressive results in machine translation, and were adopted to function on point clouds by utilizing the points as sequences. Vinyals \textit{et al}.~\cite{vinyals2015order} proposed a network that processes unordered sets using attention. They show that the network is able to sort numbers. However, they only focus on generic sets. In contrast, we present an approach that is applied to different point cloud related tasks for capturing shape and geometry information.  Recently, Lee \textit{et al}.~\cite{lee2019set} proposed Set Transformer, a method that is related to our approach. They adapt the original Transformer network to process unordered sets by using induced points, i.e. trainable parameters of the network, that are attended to the input. 
Set Transformer focuses on general sets as input. Furthermore, Lee \textit{et al}. demonstrate that it is applicable to point sets. In our work, Point Transformer is specifically designed to process point clouds and leverage important characteristics of points in metric space such as shape and geometric relations.

Xie \textit{et al}.~\cite{xie2018attentional} propose ShapeContextNet, where they hierarchically apply the shape context approach that acts as a convolutional building block. To overcome the difficulties of manually tuning the shape context parameters, Xie \textit{et al}. employ self-attention to combine the selection and feature aggregation process into one trainable operation. However, similar to point cloud convolutions, shape context relies on a manual selection of the shape context kernels which is sensitive to the irregularities of point cloud data.

The Point2Sequence model~\cite{liu2019point2sequence} uses an attention-based sequence-to-sequence network. The approach first extracts local regions and produces local features using an LSTM-based attention module. Using a set pooling method, a global feature vector is generated following the ideas of~\cite{zaheer2017deep} and~\cite{qi2017pointnet}. However, it relies on a sequence-to-sequence architecture that tends to be more computational complex than multi-head attention~\cite{vaswani2017attention}. Furthermore, in contrast to our method, Point2Sequence uses a max-pooling operation to make the network permutation invariant. 
Yang \textit{et al}.~\cite{yang2019modeling} introduce a network architecture that replaces traditional subsampling methods like furthest point sampling (FPS) with an attention-based selection process using the gumbel-softmax function, which is similar to the proposed SortNet module.

Recently, Tao et. al~\cite{9427563} proposed a multi-head attentional point cloud processing network that uses a rotation invariant representation of point clouds as input. For that, they employ a multi-head attentional convolution layer (MACL) with attention coding. However, their work focuses on designing a rotation invariant network that relies on global max pooling operations, whereas Point Transformer together with SortNet leverages the strengths and advantages of the attention operation to select useful local point structures and relates them to the global shape to induce permutation invariance.

\section{Fundamentals}\label{seq:03_fundamentals}
Attention has been first proposed for natural language processing, where the goal is to focus on a subset of important words~\cite{bahdanau2014neural}. Here, we frame the problem in the context of point sets. We consider the unordered point set \mbox{$\mathcal{P}=\{ p_i \in \mathbb{R}^{D}, i = 1, \dots, N\}$.} Our goal is to map $\mathcal{P}$ to the output space $\mathbb{R}^O$ with the set function $f:\mathcal{P} \rightarrow \mathbb{R}^O$. Furthermore, we assume that $f$ is invariant to input permutations.
Since the input point set represents some object, e.g. from laser scans, the points are not independent of each other.
We aim to make use of the attention mechanism to capture the relations between the points, as well as shape information for performing visual tasks such as object classification or segmentation. 
Next, we shortly present attention and introduce the Transformer architecture in the context of point sets.

\begin{figure*}[t]
\begin{center}
\includegraphics[width=0.95\textwidth]{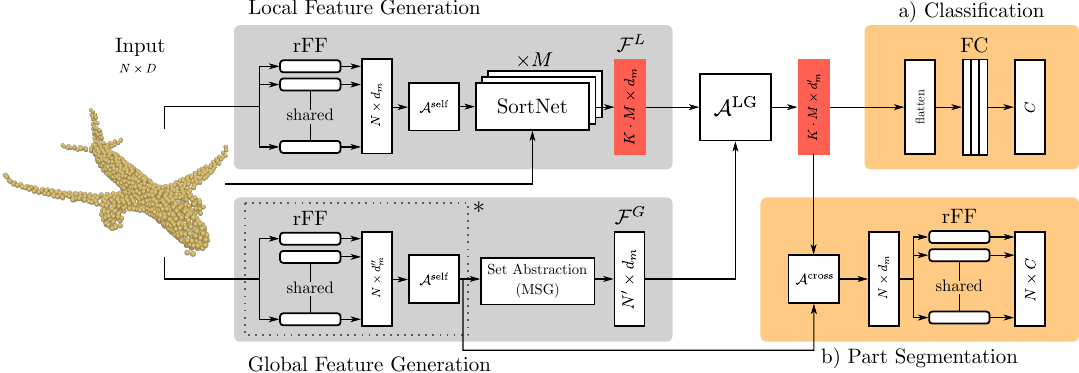}
\end{center}
\caption{Overview of the Point Transformer architecture which consists of two branches to generate local and global features. SortNet produces an ordered set of local features that are attended against the global structure of the input point cloud. Depending on the task, classification or part segmentation heads are employed.  Red Boxes denote sorted sets. * only for part segmentation.}
    \label{fig:point-transformer}
\end{figure*}

\subsection{Attention}
The idea of the attention mechanism is to set an importance-based focus on different parts of an input sequence. Consequently, relations between inputs are highlighted that can be used to capture context and higher-order dependencies. 
 The attention function $\mathcal{A}(\cdot)$ describes a mapping of $N$ queries $Q \in \mathbb{R}^{N \times d_k}$ and $N_k$ key-value pairs \mbox{$K\in \mathbb{R}^{N_k \times d_k}$,} \mbox{$V \in \mathbb{R}^{N_k \times d_v}$} to an output $ \mathbb{R}^{N\times d_k}$~\cite{vaswani2017attention}. 
Using the pairwise dot product $QK^T\in \mathbb{R}^{N\times N_k}$, a score is calculated indicating  which part of the input sequence to focus on
\begin{equation}\label{eq:attn_score}
    \text{score}(Q,K) =\sigma(QK^T),
\end{equation}
where $\text{score}(\cdot): \mathbb{R}^{N\times d_q},\mathbb{R}^{N_k\times d_k} \rightarrow \mathbb{R}^{N\times N_k}$. Furthermore, we set the activation function $\sigma(\cdot) = \text{softmax}(\cdot)$ and scale $QK^T$ by $\nicefrac{1}{\sqrt{d_k}}$ to increase stability~\cite{vaswani2017attention}. 
To capture the relations between the input points, the values $V$ are weighted by the scores from Equation~\eqref{eq:attn_score}. Therefore, we have
\begin{equation}\label{eq:attention} 
    \mathcal{A}(Q,K,V) =  \text{score}(Q,K)V,
\end{equation}
with $\mathcal{A}(Q,K,V) :  \mathbb{R}^{N\times d_k},  \mathbb{R}^{N_k\times d_k},  \mathbb{R}^{N_k\times d_v} \rightarrow \mathbb{R}^{N\times d_k}$. It is apparent, that the attention function~\eqref{eq:attention} is a weighted sum of $V$, where a value gets more weight if the dot product between the keys and values yields a higher score. 
If not specified otherwise, we set the model dimension to \mbox{$d_k = d_q = d_m$.}

\subsection{Transformer}\label{seq:transformer}
The Transformer network~\cite{vaswani2017attention} is an extension of the attention mechanism from Equation~\eqref{eq:attention} that consists of an encoder-decoder structure and introduces multi-head attention. In the following, we explain multi-head attention in detail, as our Point Transformer architecture relies on it.

Instead of employing a single attention function, multi-head attention first linearly projects the queries, keys and values $Q,K,V$ $h$ times to $d_k, d_k$ and $d_v$ dimensions, respectively, using separate feed-forward networks to learn relations from different subspaces. Then, attention is applied to each projection in parallel. The output is then concatenated and projected again using a feed-forward network. Thus, multi-head attention can be defined as follows:
\begin{equation}
    \text{Multihead}(Q,K,V) = (\text{head}_1 \oplus ... \oplus \text{head}_h) W^O,
\end{equation}
where \mbox{$\text{head}_i = \mathcal{A}(QW^Q_i, KW^ K_i, VW^V_i)$} with learnable parameters \mbox{$W^Q_i \in \mathbb{R}^{d_m \times d_k}$, $W^K_i \in \mathbb{R}^{d_m \times d_k}$} and \mbox{$W^V_i \in \mathbb{R}^{d_m \times d_v}$.}
The $\oplus$ operation denotes matrix concatenation and $W^O \in \mathbb{R}^{hd_v \times d_m}$ is a learnable parameter matrix~\cite{vaswani2017attention}. To achieve similar computational complexity as traditional attention, the dimensions of each head $d_k, d_v$ are reduced such that $d_k = d_v = d_m/h$. 
For the transformer architecture, Vaswani et al.~\cite{vaswani2017attention} define encoder and decoder stacks of identical layers that are comprised of multi-head attention and a point-wise fully connected layer, each with a residual connection followed by layer normalization~\cite{ba2016layer}. We call this layer multi-head attention and define it as follows:
\begin{equation}\label{eq:mal}
    \mathcal{A}^{\text{MH}} (X,Y) = \text{LayerNorm}(S + \text{rFF}(S)),
\end{equation}
where $\mathcal{A}^{\text{MH}} : \mathbb{R}^{N\times d_m},\mathbb{R}^{N_k \times d_m} \rightarrow  \mathbb{R}^{N\times d_m}$. 
The sublayer $S$ is defined as $S = \text{LayerNorm}(X + \text{Multihead}(X,Y,Y))$ and $\text{rFF}$ is a row-wise feed-forward network that is applied to each input independently. In practice, multiple multi-head attention layers can be deployed in sequence to further capture higher-order dependencies. Note that the output of $\mathcal{A}^{\text{MH}}$ depends on the ordering of $X$, thus it is not permutation invariant. However, the values of the corresponding outputs for each input point are always the same regardless of the input order, since $\mathcal{A}^{\text{MH}}$ only consists of matrix multiplication and summation.

For the task of point processing, we take the unordered point set $\mathcal{P}$ and generate a latent feature representation $p_i^{\text{latent}}$ with dimension $d_m$ for every $p_i \in \mathcal{P}$ using a rFF and concatenate them to form $P = [p_1^{\text{latent}}, \dots, p_N^{\text{latent}}] \in \mathbb{R}^{N\times d_m}$. Based on $P$ we now define the \textbf{self multi-head attention} as:
\begin{equation}\label{eq:smal}
    \mathcal{A}^{\text{self}} (P) := \mathcal{A}^{\text{MH}} (P,P),
\end{equation}
which performs multi-head attention between all elements of $P$, thus resulting in a matrix of same size as $P$. To attend elements of different sets, we additionally introduce a second matrix representation $Q$ of another set \mbox{$\mathcal{Q}=\{ q_j \in \mathbb{R}^{D}, j = 1, \dots, N_k\}$} that has been projected to latent feature dimension $d_m$, thus $Q \in \mathbb{R}^{N_k \times d_m}$. We can now define \textbf{cross multi-head attention} as:
\begin{equation}\label{eq:cmal}
    \mathcal{A}^{\text{cross}} (P,Q) := \mathcal{A}^{\text{MH}}(P,Q),
\end{equation}
that outputs a matrix of dimension $N\times d_m$ which order depends on the ordering of $P$. Since the output is not permutation invariant but follows the ordering of the input, Transformer and multi-head attention can not be used directly for point data without further processing. To solve this problem, we introduce our novel Point Transformer architecture that handles unordered point sets.

\section{Point Transformer}\label{seq:04_transformer}

This section presents Point Transformer, a neural network that operates on point set data and it is based on the multi-head attention mechanism. The network is permutation invariant due to a new module that we name \textbf{SortNet}. Our goal is to explore shape information of the point set by relating local and global features of the input. This is done using cross multi-head attention. To introduce our method, we first give an overview of the complete Point Transformer architecture, which is shown in Fig.~\ref{fig:point-transformer}. Our approach is divided into three parts:

\begin{enumerate}
    \item \textbf{SortNet} that extracts ordered local feature sets from different subspaces.
    \item \textbf{Global feature generation} of the whole point set.
    \item \textbf{Local-Global attention}, which relates local and global features.
\end{enumerate}

As introduced in Sec.~\ref{seq:03_fundamentals}, we consider the point set \mbox{$\mathcal{P} = \{p_i \in \mathbb{R}^D, i = 1, \dots, N\}$} as input to our network. In most cases, the point dimension is given by $D=3$ when $xyz$ coordinates are considered. Moreover, it is possible to append additional point features, for example lidar intensity values ($D=4$) or point normal vectors ($D=6$). Point Transformer consists of two independent branches: a local feature generation module, i.e. SortNet, and a global feature extraction network. For the local feature branch, the input $\mathcal{P}$ is projected to latent space with dimension $d_m$ using a row-wise feed-forward network. Then, we employ self multi-head attention on the latent features to relate the points to each other. Finally, SortNet outputs a sorted set of fixed length. This module is comparable to a kernel in convolutional neural networks, where the activation of a kernel depends on regions of the input space, i.e. the receptive field. SortNet works in a similar fashion: It focuses on points of interest according to the learnable score derived from the latent feature representation. For the extraction of global features, we employ set abstraction with multi-scale grouping introduced by~\cite{qi2017pointnet++}.
After obtaining features from both branches, we employ our proposed local-global attention to combine and aggregate local and global features of the input point cloud.
Since we use local-global attention such that the ordering of the output depends on the local features, the output of Point Transformer is permutation invariant and ordered as well and can directly be incorporated into computer vision applications such as shape classification and part segmentation.

\begin{figure*}
\begin{center}
\includegraphics[width=0.8\textwidth]{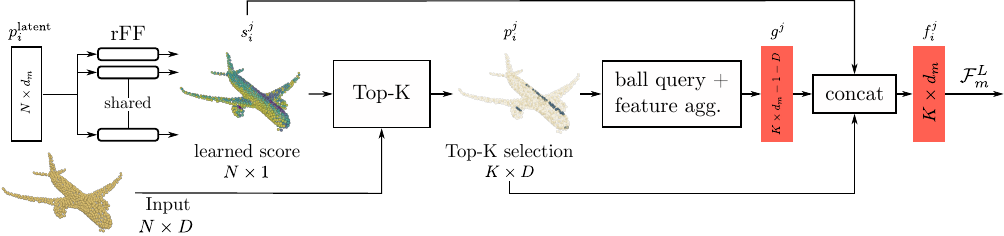}
\end{center}
\caption{Overview of the SortNet. A score is learned from a latent feature representation to extract important points from the input. Local features are aggregated from neighboring points. SortNet outputs a permutation invariant and sorted feature set. Red boxes denote sorted sets.}
    \label{fig:point-transformer_local}
\end{figure*}

\subsection{SortNet}

The local feature generation module, i.e. SortNet, is one of our key contributions. It produces local features from different subspaces that are permutation invariant by relying on a learnable score.
We show the architecture in Fig.~\ref{fig:point-transformer_local}. SortNet receives the original point cloud \mbox{$\mathcal{P} \in \mathbb{R}^{N \times D}$} and the projected latent feature representation\mbox{$P = [p_1^{\text{latent}}, \dots, p_N^{\text{latent}}] \in \mathbb{R}^{N\times d_m}$} from the row-wise feed forward network. We employ an additional self multi-head attention layer on the latent features to capture spatial and higher-order relations between each $p_i \in \mathcal{P}$.

Subsequently, a row-wise feed forward (rFF) network is used to reduce the feature dimension to one, thus creating a learnable scalar score $s_i \in \mathbb{R}$ for each input point $p_i$, which incorporates spatial relations due to the self multi-head attention layer. We now define the pair which assigns the corresponding score to every input point $\langle p_i, s_i \rangle_{i=1}^{N}$. Let $(\mathcal{Q}, \geq)$ be a totally ordered set. We select from the original input point list $K \leq N$ points with the highest score value and sort them accordingly such that:
\begin{equation}
\mathcal{Q} = \{q_j, j = 1, \dots, K\},  
\end{equation}
where $q_j = \langle p_{i}^j, s_{i}^j \rangle_{j=1}^K, p_{i}^j \in \mathcal{P}$ such that $s_{i}^1 \geq \dotsc \geq s_{i}^K$.  In other words, we employ the top-k operation to search for the $K$ highest scores $s_i$ and select the associated input points $p_i$.
After selecting $K$ points using the learnable score, we now capture localities by grouping all points from $\mathcal{P}$ that are within the euclidean distance $r$ of each selected points, i.e. we perform a ball query search similar to~\cite{qi2017pointnet++}. The grouped points are then used to encode local features, denoted by $g^j \in \mathbb{R}^{d_m - 1 - D}, j = 1, \dots, K$. We choose the feature dimension of the grouped points $g^j$ such that the resulting dimension of the local feature vector corresponds to the model dimension $d_{m}$.
The scores $s_{i}^j$, as well as the local features $g^j$ from the grouping layer, are concatenated to the corresponding input points $p_{i}^j$ to include the score calculation into our optimization problem and encode local characteristics to the selected point. Thus, we obtain our local feature vector 
\begin{equation}
    f_{i}^j = p_{i}^j \oplus s_{i}^j \oplus g^j , \quad f_{i}^j \in \mathbb{R}^{d_{m}}.
\end{equation}
Consequently, the output of SortNet constitutes one local feature set
\begin{equation}
\mathcal{F}_{m}^L = \{ f_{i}^j, j = 1,\dots, K \}.    
\end{equation}
Since $\mathcal{Q}$ is an ordered set, it follows that $\mathcal{F}_{m}^{L}$ is ordered as well.
To capture dependencies and local features from different subspaces, we employ $M$ separate SortNets. Finally, the $M$ feature sets are concatenated to obtain an ordered local feature set of fixed size
\begin{equation}
    \mathcal{F}^{L} = \mathcal{F}_{1}^L \cup \dotsc \cup \mathcal{F}_{M}^L,\quad \mathcal{F}^{L}  \in \mathbb{R}^{K\cdot M\times d_{m}}.
\end{equation}

\subsection{Global Feature Generation}
The second branch of Point Transformer is responsible for extracting global features from the input point cloud. 
To reduce the total number of points to save computational time and memory, we employ the set abstraction multi-scale grouping (MSG) layer introduced by Qi et al.~\cite{qi2017pointnet++}. We subsample the entire point cloud to $N' < N$ points using the furthest point sampling algorithm (FPS) and find neighboring points to aggregate features of dimension $d_m$ resulting in a global representation of dimension $N' \times d_{m}$. 
Note that the global feature representation is still unordered since no sorting or set pooling operation was performed. 

\subsection{Local-Global Attention}
The goal of Point Transformer is to relate local and global feature sets, $\mathcal{F}^{L} $ and $\mathcal{F}^{G}$ respectively, to capture shape and context information of the point cloud.
After obtaining both feature lists, we employ self multi-head attention $\mathcal{A}^{\text{self}}$ on the local features $\mathcal{F}^L$ as well as the global features $\mathcal{F}^G$.
Then, cross multi-head attention layer $\mathcal{A}^{\text{cross}}$ from Equation \eqref{eq:cmal} is applied such that every global feature is scored against every local feature, thus relating local context with the underlying shape. We call this operation local-global attention $\mathcal{A}^{\text{LG}}$ (see Fig.~\ref{fig:point-transformer}) and define it as follows:
\begin{equation}
    \mathcal{A}^{\text{LG}} := \mathcal{A}^{\text{cross}} ( \mathcal{A}^{\text{self}} (F^L) , \mathcal{A}^{\text{self}} (F^G) ),
\end{equation}
where $F^L$ and $F^G$ are the matrix representations of $\mathcal{F}^L$ and $\mathcal{F}^G$, respectively. The last row-wise feed forward layer in the multi-head attention mechanism of $\mathcal{A}^{\text{LG}}$ reduces the feature dimension to $d_m' < d_m$ in order to decrease computational complexity, thus we have 
\mbox{$\mathcal{A}^{\text{LG}} : \mathbb{R}^{K\cdot M\times d_{m}}, \mathbb{R}^{N'\times d_{m}} \rightarrow \mathbb{R}^{K\cdot M\times d'_{m}}$.} 
In other words, we take every local feature from SortNet and score the global features against it. 
At this point, it is important to note that we relate the local features, i.e. a subset of the input $\mathcal{F}^L \subseteq \mathcal{P}$, with the global structure. Thus, we avoid reducing the shape representation using set pooling; instead, the output of local-global attention includes information of the entire point cloud, i.e. the underlying shape, as well as local characteristics.
As with multi-head attention, for local-global attention, we employ multiple cross and self multi-head attention layers in sequence to learn higher-order dependencies~\cite{vaswani2017attention}. Since the ordering of the local features $\mathcal{F}^L$ defines the order of the output of local-global attention, we obtain a permutation invariant latent representation of fixed size of the aggregated features, that can directly be incorporated into computer vision tasks.

\subsection{Complete Model}
\begin{figure*}
\begin{center}
\includegraphics[width=\textwidth]{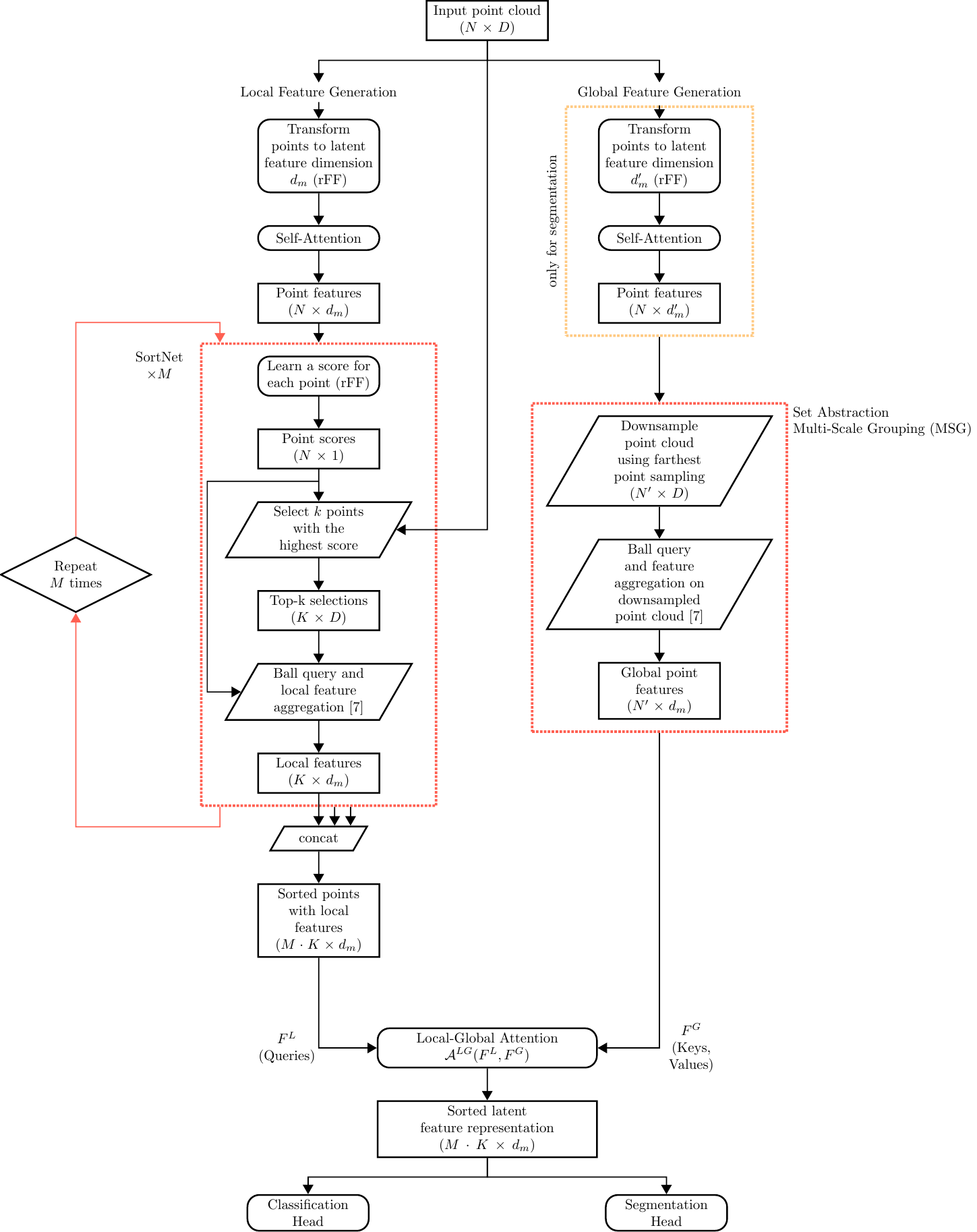}
\end{center}
\caption{Overview of the processing chain of Point Transformer. Data is shown as rectangles with the respective dimensions. Networks modules, for example row-wise feed forward networks (rFF), are denoted by rectangles with rounded corners and additional process steps are shown as parallelograms. Here, it is important to note that individual rFF's  with separate weights are deployed in each of the $M$ SortNet modules.}
    \label{fig:point-transformer_flowchart}
\end{figure*}

To recap, Point Transformer functions as follows: Our architecture is comprised of two independent branches, SortNet for the extraction of local features and a global feature generation module. SortNet constitutes a novel architecture that selects a number of input points based on a learned score from latent features, resulting in $M \cdot K$ ordered feature vectors with dimension $d_{\text{m}}$. In the global feature branch, we employ multi-scale grouping to reduce the total number of points to $N'$ while aggregating spatial information. Then, local-global attention is used to relate both spatial signatures, producing a permutation invariant and ordered representation of length $K \cdot M$ with reduced dimension $d'_m$ (see Fig.~\ref{fig:point-transformer}), which can be used for different tasks such as shape classification or part segmentation. Additionally, we demonstrate the processing chain of our model as a flowchart in Fig.~\ref{fig:point-transformer_flowchart}.

\textbf{Shape Classification} assigns the point cloud to one of $C$ object classes. For this, we flatten the sorted output of local-global attention to a vector of fixed size $\mathbb{R}^{M\cdot K \cdot d'_m}$ and reduce the dimensions using a row-wise feed-forward network to $\mathbb{R}^C$. Thus, each output represents one class. Using a final softmax layer, class probabilities are produced. The shape classification head is shown in Fig.~\ref{fig:point-transformer} a). 

\textbf{Part Segmentation} assigns a label to each point of the input set. State-of-the-art methods~\cite{qi2017pointnet++,liu2019point2sequence} upsample a global feature vector obtained from a set pooling operation using interpolation. We, however, employ an additional cross multi-head attention layer to attend the output of $\mathcal{A}^{\text{LG}}$, i.e. the aggregated shape and context information, to each point of the input set $\mathcal{P}$. 
It is important to note that we project the points in the global feature generation branch to $d_m''$ dimensions and apply self multi-head attention. The features are additionally used for the set abstraction layer. Later, we attend the projected features with the output of Point Transformer. Thus, we can relate each point to the entire point cloud.
The result is a matrix of dimension $\mathbb{R}^{N \times d'_{m}}$. Then, a row-wise feed-forward layer reduces the dimension of each point to the $C$ possible classes $\mathbb{R}^{N\times C}$. Again, using a final softmax layer, per-point class probabilities are produced as shown in Fig.~\ref{fig:point-transformer} b).

\section{Experiments}\label{seq:05_exp}

\begin{table}[!t]
    \centering
        \caption{Here, we compare Point Transformer to related approaches that use either set pooling or attention. We evaluate on popular Benchmarks for object classification (ModelNet) and part segmentation (ShapeNet). }  
        \label{tab:modelnet40_results} 
        \begin{tabular}{@{}lcc@{}}
        \toprule
        Method             & ModelNet & ShapeNet \\ \midrule
        PointNet~\cite{qi2017pointnet}        & 89.2 &    83.7                          \\
        PointNet++~\cite{qi2017pointnet++}     & 91.9 &  85.1                            \\
         ShapeContextNet~\cite{xie2018attentional} & 89.8 &  84.6
            \\
        Deep Sets~\cite{zaheer2017deep}         & 90.3 &  -                           \\
       
        Point2Sequence~\cite{liu2019point2sequence} & 92.6 & 85.2                         \\
        Set Transformer~\cite{lee2019set}    & 90.4 &   -                           \\ 
        PAT~\cite{yang2019modeling}& 91.7 & - \\
        Tao et. al~\cite{9427563} & 87.5 & 75.2 \\
        Point Transformer & 92.8 & 85.9
        \\\midrule
        KPConv~\cite{thomas2019kpconv} & 92.9 &  86.2 \\
        PointCNN~\cite{li2018pointcnn} & 92.2 & 86.1 \\
        SpiderCNN~\cite{xu2018spidercnn} & 90.5 & 85.3 \\
        \bottomrule
        \end{tabular}
 \end{table}

In this section, we perform two standard evaluations on Point Transformer. We compare our results with approaches that operate directly on 3D point sets~\cite{qi2017pointnet,qi2017pointnet++,zaheer2017deep}, attention-based approaches~\cite{xie2018attentional,lee2019set,liu2019point2sequence} and methods that use point cloud convolutions
~\cite{thomas2019kpconv,li2018pointcnn,wang2019dynamic,xu2018spidercnn}. Moreover, we provide a thoughtful analysis and visualizations of the components of our approach. We implement our network in Pytorch~\cite{paszke2017automatic} where we rely on the RAdam optimizer~\cite{liu2019variance} for all experiments. The weights of each layer are initialized using the popular Kaiming normal initialization method~\cite{he2015delving}. Our implementation will be made publicly available.

\subsection{Point cloud classification}

We evaluate Point Transformer on the ModelNet40 dataset~\cite{wu20153d} and use the modified version by Qi \textit{et al}.~\cite{qi2017pointnet++} that provides $10.000$ points sampled from the mesh of the CAD model, as well as the normal vectors for each point. The dataset consists of $40$ categories and it is composed of $9843$ training samples and $2468$ test samples. During the training for classification, we augment the input by randomly scaling the shape in the range of $[0.8, 1.25]$ and randomly translating in the range of $[-0.1, 0.1]$. Additionally, we apply random dropout of the input points as proposed in~\cite{qi2017pointnet,qi2017pointnet++}. For the experiments, we set $N=1024$, $D = 6$ ($xyz$ and normals), $d_m = 512$, $d'_m = 64$, $M=4$ and $K=64$.
The results of the shape classification are shown in Table~\ref{tab:modelnet40_results}. Point Transformer outperforms attention-based methods (top part of Table~\ref{tab:modelnet40_results}) and achieves on par accuracy when compared to state-of-the art methods (bottom part of Table~\ref{tab:modelnet40_results}) with a classification accuracy of $92.8\%$.

\begin{table}[!t]
    \centering
        \caption{Results of Network Design Analysis. We evaluate different SortNet architectures to highlight that the learnable score increases the networks performance. Additionally, we compare different sampling methods for the global feature generation branch.}
        \label{tab:ablation_study_1}
        \begin{tabular}{@{}lc@{}}
            \toprule
            a) Ablation Study SortNet & \multicolumn{1}{l}{Accuracy (\%)} \\ \midrule
            SortNet with learnable score  & 83.4 \\
            SortNet with FPS & 74.8 \\
            SortNet with random points  & 60.1 \\
            \toprule
            b) Ablation Study Global  & \\
            Feature Generation & Accuracy (\%) \\ \midrule
            No sampling & 91.9 \\
            FPS ($N'=128$) & 92.3 \\
            Set Abstraction (MSG) ($N'=128$) & 92.8 \\
            \bottomrule
        \end{tabular}
 \end{table}
 
\subsection{Point cloud part segmentation}
\begin{figure*}[]
    \centering
    \includegraphics[height=0.97\textheight]{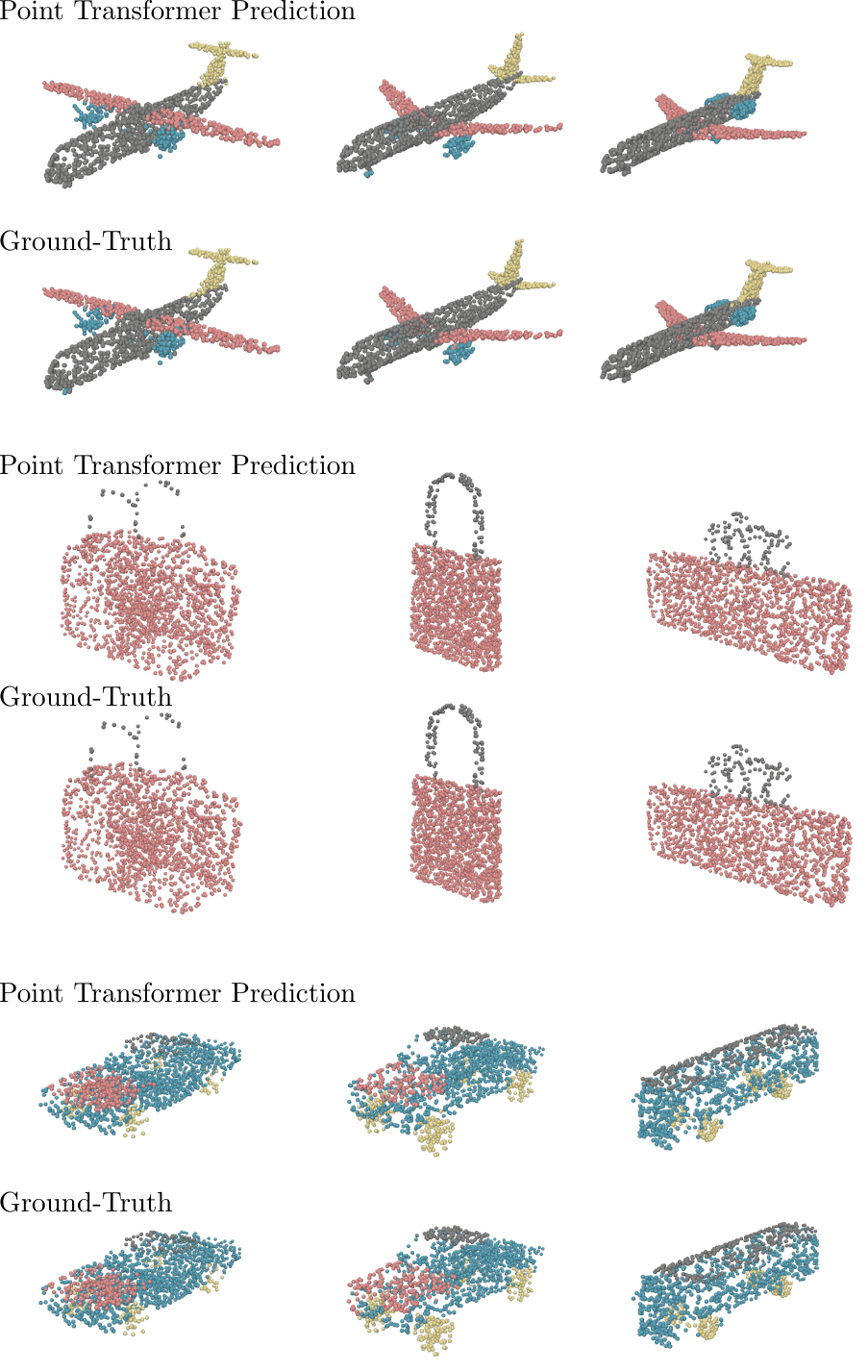}
    \caption{Additional results of the part segmentation task for different object categories. We show the prediction of Point Transformer (top) in comparison with the ground-truth (bottom).}
    \label{fig:part_seg_1}
\end{figure*}

Here, we evaluate Point Transformer on the challenging task of point cloud part segmentation on the ShapeNet dataset~\cite{shapenet2015}, which contains $13.998$ train samples and $2874$ test samples. The dataset is composed of objects from $16$ categories with a total of $50$ part labels. The goal is to predict the class category of every point. To address this task, the network has to learn a deep understanding of the underlying shape. For the part segmentation, we set $M=10$ and $K=16$. Again, we use $xyz$ coordinates with normal vectors ($D=6$) and $N=1024$ input points. For this experiment, we follow the setup of~\cite{qi2017pointnet} where a one-hot encoding of the  category is concatenated to the input points as an additional feature. We report the mean IoU (Intersection-over-Union) in Table~\ref{tab:modelnet40_results}. Finally, we visualize exemplary results of the part segmentation task in Fig.~\ref{fig:part_seg_1}.

\subsection{Network complexity}
We examine the network complexity of Point Transformer and perform a comparison to related approaches. The results of this experiment are shown in Table~\ref{tab:model_size}. We performed all experiments on a Nvidia GeForce 1080Ti. Point Transformer has about $13.5$ million learnable parameters ($51$~MB), which is less when compared to KPConv ($15$ million learnable parameters). However, our model is about $6$ times bigger than PointNet++ and Point2Seq. This is mainly due to the fact that the Transformer model itself has a lot of learnable parameters. For example, one SortNet only has about $10.000$ learnable parameters which shows that SortNet can be incorporated into any existing network architecture without much space requirements and computational overhead, as it only adds about $1.2$~ms of inference time. In many cases, the forward pass of multiple SortNets can additionally be performed in parallel. Even though, Point Transformer has more learnable parameters than, e.g, PointNet++, it still has a faster inference time because multi-head attention blocks are highly optimized and computation is also performed in parallel by employing multiple attention heads. For the computational complexity of the network, an upper bound can be estimated from the most expensive operation, which in our case is the multi-head attention mechanism. The complexity is given by $\mathcal{O}(N^2\cdot d_m)$, thus it scales quadratic with respect to the total number of input points.

\begin{table}[t!]
\centering
\caption {Model complexity study. Here, we compare the network size and the inference time against related approaches.} \label{tab:model_size} 
\begin{tabular}{@{}llll@{}}
\toprule
Method            & \# of params & Network size & Inference time \\ \midrule
Point Transformer & 13.5 M & 51 MB        & 110 ms   \\
SortNet           & 10 k         &   0.04 MB           & 1.25 ms    \\ \midrule
PAT~\cite{yang2019modeling} & - & 5.8 MB & 88 ms \\ 
KPConv~\cite{thomas2019kpconv}           & 15 M   & -            & 210 ms   \\
PointNet++~\cite{qi2017pointnet++}        & 2.1 M        & 24 MB        & 160 ms   \\
Point2Seq~\cite{liu2019point2sequence}         & 2 M          & -            & -              \\ \bottomrule
\end{tabular}
\end{table}

\subsection{Hyperparameter Study}
Here, we analyze the effects of different numbers of SortNets in our Point Transformer architecture as well as the amount of Top-K selections on the ModelNet40 dataset~\cite{wu20153d}. The results are shown in Tab.~\ref{tab:hyperparams}. Furthermore, we present the hyperparameters that were used for the reported results for the classification and the part segmentation task in Tab.~\ref{tab:hyperparams_network}. The parameters follow the notation introduced in Fig.~\ref{fig:point-transformer} and Fig.~\ref{fig:point-transformer_local}. The values were found by performing a hyperparameter grid search experiment for the classification and the part segmentation, similar to Tab.~\ref{tab:hyperparams}. We report the set of parameters that achieved the best overall performance. Note, that for the rFF, each value in the parenthesis denotes one layer, where the value represents the feature dimension for that layer.

\begin{table*}[th!]
 \centering
    \caption {Hyperparameter Study Results on ModelNet40 for different combinations of the hyperparameters M (number of SortNets) and K (Top-K selections).} \label{tab:hyperparams} 
    \resizebox{\textwidth}{!}{%
\begin{tabular}{@{}lcccccccccccccccc@{}}
\toprule
\textbf{M}        & 4    & 4    & 4             & 4    & 8    & 8    & 8    & 8    & 8    & 16   & 16   & 16   & 16   & 32   & 32   & 32   \\
\textbf{K}        & 16   & 32   & 64            & 96   & 8    & 16   & 32   & 64   & 96   & 8    & 16   & 32   & 64   & 8    & 16   & 32   \\ \midrule
\textbf{Accuracy} & 91.7 & 92.3 & \textbf{92.8} & 91.7 & 90.2 & 90.5 & 91.9 & 92.4 & 92.0 & 91.2 & 91.6 & 92.0 & 91.7 & 90.8 & 91.3 & 91.1 \\ \bottomrule
\end{tabular}
}
\end{table*}

\begin{table}[th!]
 \centering
    \caption {Hyperparameters of Point Transformer for the classification and the part segmentation task.} \label{tab:hyperparams_network} 
 \resizebox{\columnwidth}{!}{%
\begin{tabular}{@{}llcc@{}}
\toprule
Parameter & Explanation & Classification & Part Segmentation\\ \midrule
\multicolumn{4}{l}{\textbf{General Hyperparameters}} \\
B & Batch size & $11$ & $8$ \\
N & Number of input points & $1024$ & $1024$ \\
D & Input dimension & $6$ & $6$ \\ 
$l_r$ & Learning rate & $0.001$ & $0.005$ \\
$w_d$ & Weight decay & $\num{1e-6}$ & $0.0001$ \\
$d_m$ & Latent feature dimension & $512$ & $512$ \\
- & Weight initializer & \multicolumn{2}{c}{Kaiming Normal~\cite{he2015delving}} \\
\multicolumn{4}{l}{\textbf{Local Feature Generation}} \\
- & rFF feature dimension & $(64, 128, 512)$ & $(64, 128, 512)$ \\
- & Dropout & $0.4$ & $0.3$ \\
$n_\text{head}$ & Number of local attention heads & $8$ & $8$ \\
$n_\text{layers}$ & Number of local attention layers & $1$ & $1$ \\
\multicolumn{4}{l}{\textbf{SortNet}} \\
$M$ & Number of SortNets  & $4$ & $10$ \\
$K$ & Number of top-k selections & $64$ & $16$ \\
- & rFF feature dimension & $(128, 256, 512)$ & $(64, 128, 512)$ \\
- & Dropout & $0.4$ & $0.3$ \\
\multicolumn{4}{l}{\textbf{Global Feature Generation}} \\
$N'$ & Reduced point set & $128$ & $64$ \\
- & Segmentation rFF & - & $(64, 128, 256)$ \\
$d_m^{''}$ & Segmentation feature dimension & - & $256$ \\
$n_\text{head}$ & Number of local attention heads & $8$ & $8$ \\
$n_\text{layers}$ & Number of local attention layers & $1$ & $1$ \\
- & Dropout & $0.4$ & $0.3$ \\
\multicolumn{4}{l}{\textbf{Local-Global Attention}} \\
$d_m^{'}$ & Reduced feature dimension & $64$ & $256$ \\
$n_\text{head}$ & Number of local attention heads & $8$ & $8$ \\
$n_\text{layers}$ & Number of local attention layers & $4$ & $4$ \\
\multicolumn{4}{l}{\textbf{Classification Head}} \\
C & Number of classes & \multicolumn{2}{c}{$40$} \\
- & Fully connected dimension& \multicolumn{2}{c}{$(4096, 1024, 512, 128, 40)$} \\
- & Dropout & \multicolumn{2}{c}{$0.4$} \\
\multicolumn{4}{l}{\textbf{Segmentation Head}} \\
C & Number of classes & \multicolumn{2}{c}{$50$} \\
- & Output rFF& \multicolumn{2}{c}{$(256, 128, 50)$} \\
$n_\text{head}$ & Number of local attention heads & $8$ & $8$ \\
$n_\text{layers}$ & Number of local attention layers & $1$ & $1$ \\
 \bottomrule
 
\end{tabular}
}

\end{table}

\subsection{Point Transformer Design Analysis}
We conduct an ablation study to show the influence of each Point Transformer module. Afterward, we qualitatively examine our classification results by visualizing the learned point set regions that contribute to the classification output. 

\textbf{Ablation study of SortNet:}
We first evaluate Point Transformer using only the SortNet module from Fig.~\ref{fig:point-transformer_local} with the classification head from Fig.~\ref{fig:point-transformer} a). Our aim is to show that the learned scores are based on the importance of points for the classification task. In addition, we want to verify that SortNet selects points that help to understand the underlying shape. Since we cannot explicitly define which are the most important points, we rely on the accuracy score.
In detail, we train SortNet based on three different experiments and deliberately set $M=10$ and $K=12$, selecting only a subset of the entire point cloud ($M\cdot K = 120$, $N = 1024$).
In the first experiment, we train SortNet as it is implemented in the Point Transformer pipeline. 
In the second experiment, we replace the Top-K selection process with the furthest point sampling.
Finally, we randomly select $K$ points from the input set instead of the learned Top-K selection. It is important to note, that the last two experiments remove the permutation invariance property. However, we want to show that SortNet performs better than a random selection of points and handcrafted sampling methods. Thus, we rely on random sampling and FPS as baselines.
The results are shown in Table~\ref{tab:ablation_study_1} a). With randomly sampled points, SortNet achieves $60.1\%$ classification accuracy. When we apply the FPS to cover most of the underlying shape, the accuracy increases to $74.8\%$, indicating  spatial information preservation. Finally, when we use learned Top-K selection, we achieve the highest classification accuracy of $83.4\%$. This empirically shows that SortNet learns to focus on important shape regions.

\textbf{Ablation study Global Feature Generation:}
In this ablation study, we compare different sampling methods for the extraction of global features. We rely on the complete Point Transformer pipeline as shown in Fig.~\ref{fig:point-transformer} and replace the set abstraction (MSG) with different sampling approaches.
Again, we evaluate the accuracy of the classification task. The results are presented in Table~\ref{tab:ablation_study_1} b). In the first experiment, we use the complete input point cloud. Then, we sample $N'=128$ points using the furthest point sampling, which slightly improves our result by $0.4\%$. When we additionally aggregate features from local regions around the sampled points, i.e. set abstraction with multiscale grouping (MSG)~\cite{qi2017pointnet++}, the accuracy can be further increased to $92.8\%$. 
This indicates that scoring the local features against every input point makes it harder to find important relations. Additionally, by uniformly selecting fewer points and aggregating local features the network can concentrate on meaningful parts of the underlying shape.

\begin{figure*}[]
    \centering
    \includegraphics[width=0.97\textwidth]{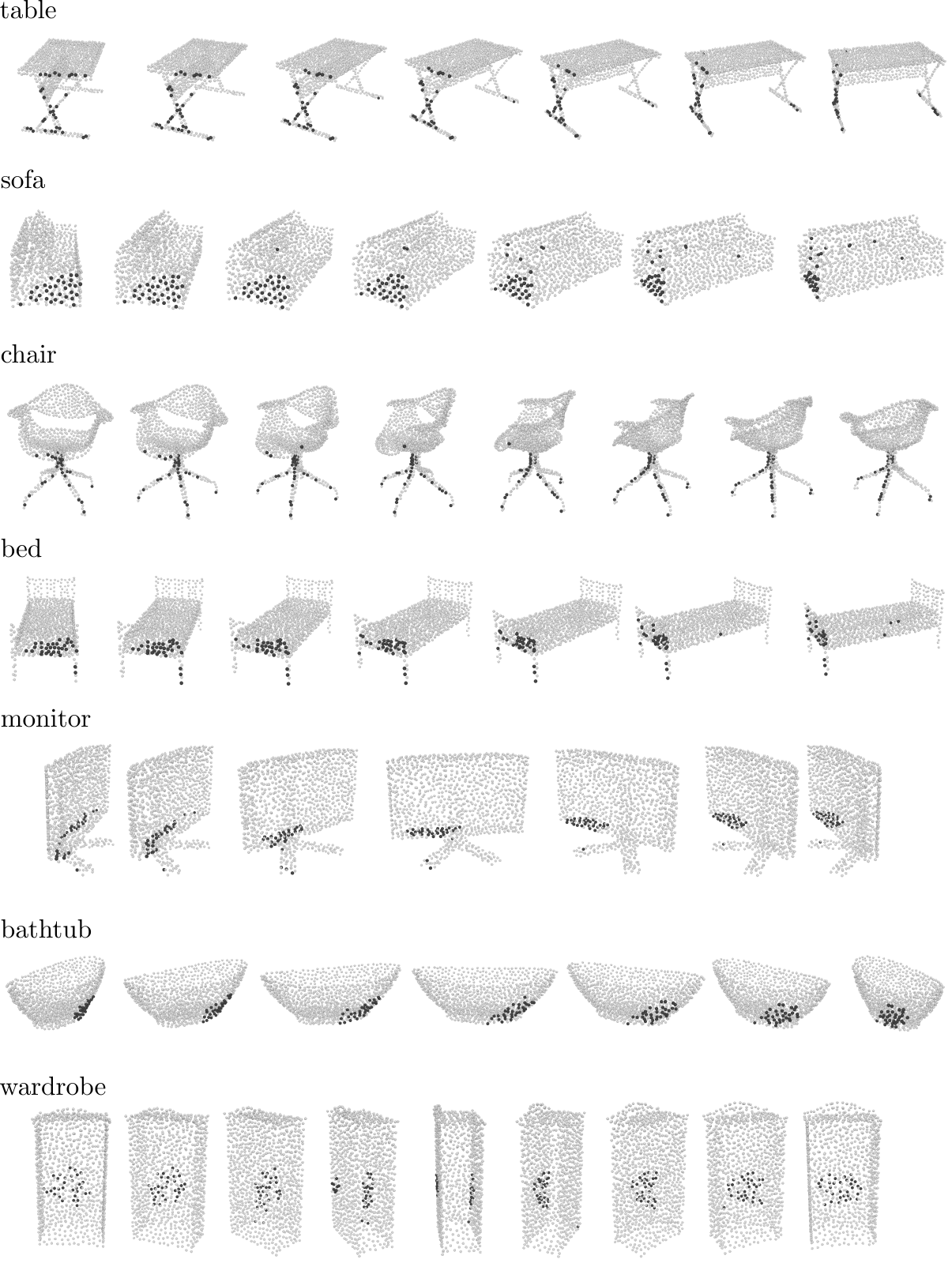}
    \caption{Influence of input point rotations on the Top-K selection process.  \textcolor{color_selection}{\textbullet} Top-K selection, \textcolor{color_pcl}{\textbullet} Input points. When the input point cloud is rotated, SortNet still focuses on similar local regions of the underlying shape.}
    \label{fig:rot_1}
\end{figure*}

\textbf{Rotation robustness of SortNet:}
In this section we evaluate the robustness of SortNet against rotations of the input cloud. For this, we first evaluate Point Transformer on the ModelNet40 test set and randomly rotate the input point cloud. Even though we did not train the network with rotations, we still achieve a classification accuracy of $92.3\%$ compared to $92.8\%$ without rotations. We applied the same input point rotation to PointNet++ and classification accuracy dropped from $91.9\%$ to $88.6\%$. To qualitatively support this claim, we visualize the learned Top-K selections of one SortNet for different rotations in Fig.~\ref{fig:rot_1}, which shows that SortNet still focuses on the similar local regions even when the input point cloud is rotated.

\begin{figure*}[]
    \centering
    \includegraphics[width=0.85\textwidth]{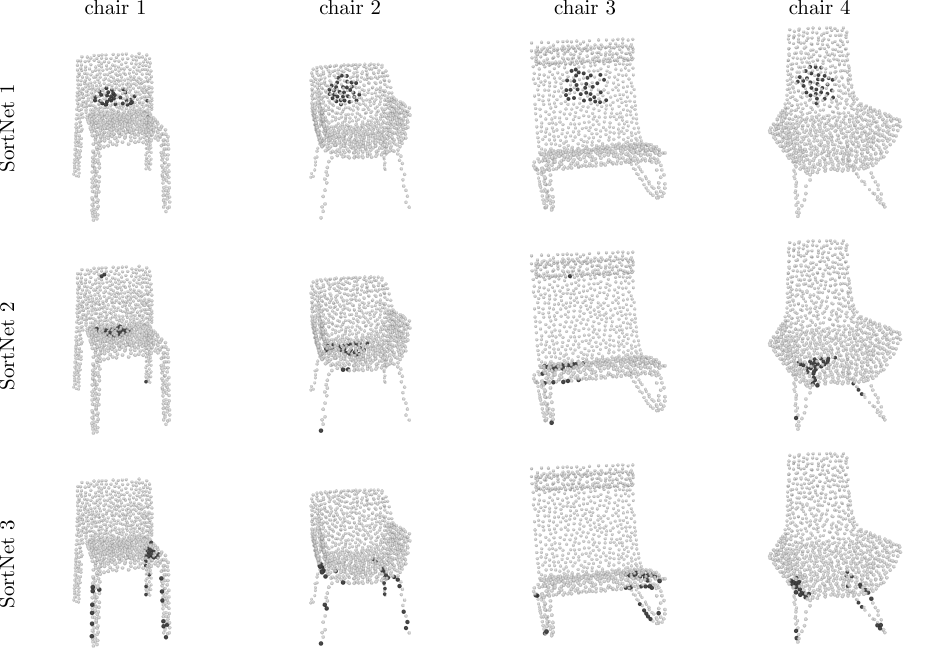}
    \caption{Top-K selections for different chair models.  \textcolor{color_selection}{\textbullet} Top-K selection (dark points), \textcolor{color_pcl}{\textbullet} Input points (light points). SortNet selects points from similar local regions when applied to objects of the same category, suggesting that it is aware of the underlying shape.}
    \label{fig:kernel_chair}
\end{figure*}

\begin{figure*}[]
    \centering
    \includegraphics[width=0.85\textwidth]{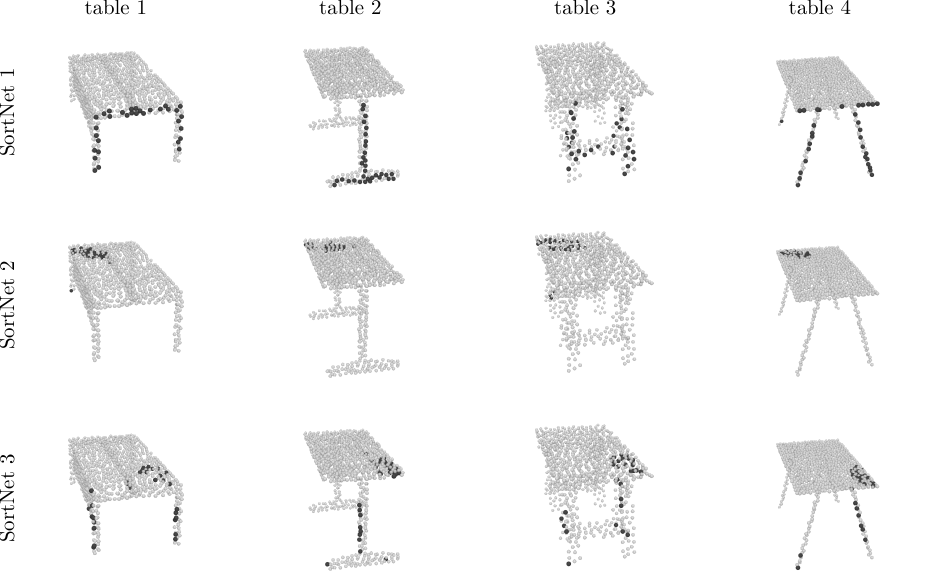}
    \caption{Top-K selections for different table models.  \textcolor{color_selection}{\textbullet} Top-K selection, \textcolor{color_pcl}{\textbullet} Input points. SortNet selects points from similar local regions when applied to objects of the same category, suggesting that it is aware of the underlying shape.}
    \label{fig:kernel_table}
\end{figure*}

\textbf{Visualizations of learned local regions:} 
Here, we show that SortNet focuses on local regions similar to the receptive field of a CNN. For this, we visualize the learned Top-K selections of multiple trained SortNet modules on different models of the same object class in Fig.~\ref{fig:kernel_chair} and Fig.~\ref{fig:kernel_table}. It is apparent, that each SortNet tries to select similar regions even when the shape of the model is slightly different. This, together with the results from the rotational robustness, suggests that SortNet is aware of the underlying shape.

\begin{figure*}[]
    \centering
    \includegraphics[width=\textwidth]{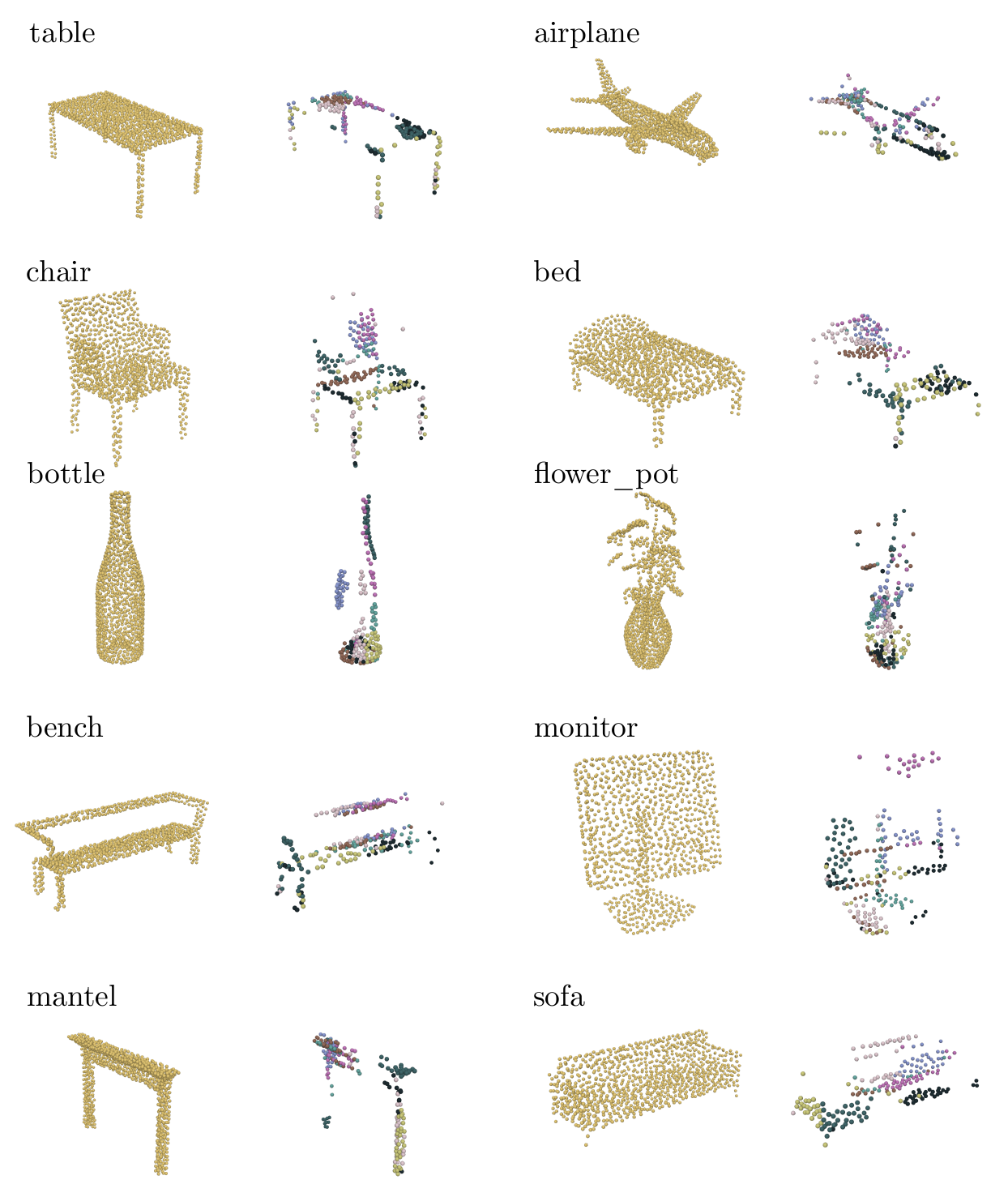}
    \caption{Here all selected points  from the local feature generation branch (right) are shown in comparison with the complete input point cloud (left). The selected points of each SortNet are shown in the same color. It is clear that every SortNet focuses on different local regions of the object. When the selected points are visualized together, the input point cloud is still recognizable, suggesting that in combination,  all SortNets try to retain as much as possible of the underlying shape.}
    \label{fig:selection}
\end{figure*}

\textbf{All Top-K selections:} 
As an additional evaluation, we show all selected points of $M=8$ SortNet modules in Fig.~\ref{fig:selection} for the classification task. We visualize points that were selected from the same SortNet with the same color. It is apparent, that different SortNet modules focus on different parts of the object and in combination, still retain as much as possible of the underlying shape.

\section{Conclusion and Future Work}\label{seq:06_conclusion}
In this work, we proposed Point Transformer, a permutation invariant neural network that relies on the multi-head attention mechanism and operates on irregular point clouds. The core of Point Transformer is a novel module that receives a latent feature representation of the input point cloud and selects points based on a learned score. We relate local features to the global structure of the point cloud, thus exploiting context and inducing shape-awareness. The output of Point Transformer is a sorted and permutation invariant feature list that is used for shape classification and part segmentation. Finally, we show that our point selection mechanism is based on importance for the specified task.
As future work, we want to focus on improving the efficiency of the Transformer architecture by implementing recent advances for self-attention, such as~\cite{wang2020linformer, xiong2021nystr}.

\EOD


\begin{thebibliography}{10}

\bibitem{lang2019pointpillars}
Alex~H Lang, Sourabh Vora, Holger Caesar, Lubing Zhou, Jiong Yang, and Oscar
  Beijbom.
\newblock Pointpillars: Fast encoders for object detection from point clouds.
\newblock In {\em Proceedings of the IEEE Conference on Computer Vision and
  Pattern Recognition}, pages 12697--12705, 2019.

\bibitem{engel2018deep}
Nico Engel, Stefan Hoermann, Philipp Henzler, and Klaus Dietmayer.
\newblock Deep object tracking on dynamic occupancy grid maps using rnns.
\newblock In {\em 2018 21st International Conference on Intelligent
  Transportation Systems (ITSC)}, pages 3852--3858. IEEE, 2018.

\bibitem{engel2019deeplocalization}
Nico Engel, Stefan Hoermann, Markus Horn, Vasileios Belagiannis, and Klaus
  Dietmayer.
\newblock Deeplocalization: Landmark-based self-localization with deep neural
  networks.
\newblock In {\em 2019 IEEE Intelligent Transportation Systems Conference
  (ITSC)}, pages 926--933. IEEE, 2019.

\bibitem{simon2020stickypillars}
Martin Simon, Kai Fischer, Stefan Milz, Christian~Tobias Witt, and
  Horst-Michael Gross.
\newblock Stickypillars: Robust feature matching on point clouds using graph
  neural networks.
\newblock {\em arXiv preprint arXiv:2002.03983}, 2020.

\bibitem{horn2020deepclr}
Markus Horn, Nico Engel, Vasileios Belagiannis, Michael Buchholz, and Klaus
  Dietmayer.
\newblock Deepclr: Correspondence-less architecture for deep end-to-end point
  cloud registration.
\newblock In {\em 2020 IEEE 23rd International Conference on Intelligent
  Transportation Systems (ITSC)}, pages 1--7. IEEE, 2020.

\bibitem{wiederer2020}
Julian Wiederer, Arij Bouazizi, Ulrich Kressel, and Vasileios Belagiannis.
\newblock Traffic control gesture recognition for autonomous vehicles.
\newblock In {\em 2020 IEEE/RSJ International Conference on Intelligent Robots
  and Systems (IROS)}, pages 10676--10683, 2020.

\bibitem{qi2017pointnet++}
Charles~Ruizhongtai Qi, Li~Yi, Hao Su, and Leonidas~J Guibas.
\newblock Pointnet++: Deep hierarchical feature learning on point sets in a
  metric space.
\newblock In {\em Advances in neural information processing systems}, pages
  5099--5108, 2017.

\bibitem{qi2019deep}
Charles~R Qi, Or~Litany, Kaiming He, and Leonidas~J Guibas.
\newblock Deep hough voting for 3d object detection in point clouds.
\newblock In {\em Proceedings of the IEEE International Conference on Computer
  Vision}, pages 9277--9286, 2019.

\bibitem{maturana2015voxnet}
Daniel Maturana and Sebastian Scherer.
\newblock Voxnet: A 3d convolutional neural network for real-time object
  recognition.
\newblock In {\em 2015 IEEE/RSJ International Conference on Intelligent Robots
  and Systems (IROS)}, pages 922--928. IEEE, 2015.

\bibitem{wu20153d}
Zhirong Wu, Shuran Song, Aditya Khosla, Fisher Yu, Linguang Zhang, Xiaoou Tang,
  and Jianxiong Xiao.
\newblock 3d shapenets: A deep representation for volumetric shapes.
\newblock In {\em Proceedings of the IEEE conference on computer vision and
  pattern recognition}, pages 1912--1920, 2015.

\bibitem{qi2016volumetric}
Charles~R Qi, Hao Su, Matthias Nie{\ss}ner, Angela Dai, Mengyuan Yan, and
  Leonidas~J Guibas.
\newblock Volumetric and multi-view cnns for object classification on 3d data.
\newblock In {\em Proceedings of the IEEE conference on computer vision and
  pattern recognition}, pages 5648--5656, 2016.

\bibitem{su2015multi}
Hang Su, Subhransu Maji, Evangelos Kalogerakis, and Erik Learned-Miller.
\newblock Multi-view convolutional neural networks for 3d shape recognition.
\newblock In {\em Proceedings of the IEEE international conference on computer
  vision}, pages 945--953, 2015.

\bibitem{qi2017pointnet}
Charles~R Qi, Hao Su, Kaichun Mo, and Leonidas~J Guibas.
\newblock Pointnet: Deep learning on point sets for 3d classification and
  segmentation.
\newblock In {\em Proceedings of the IEEE Conference on Computer Vision and
  Pattern Recognition}, pages 652--660, 2017.

\bibitem{zaheer2017deep}
Manzil Zaheer, Satwik Kottur, Siamak Ravanbakhsh, Barnabas Poczos, Ruslan~R
  Salakhutdinov, and Alexander~J Smola.
\newblock Deep sets.
\newblock In {\em Advances in neural information processing systems}, pages
  3391--3401, 2017.

\bibitem{wagstaff2019limitations}
Edward Wagstaff, Fabian Fuchs, Martin Engelcke, Ingmar Posner, and Michael~A.
  Osborne.
\newblock On the limitations of representing functions on sets.
\newblock In {\em Proceedings of the 36th International Conference on Machine
  Learning, {ICML} 2019}, volume~97 of {\em Proceedings of Machine Learning
  Research}, pages 6487--6494. {PMLR}, 2019.

\bibitem{bahdanau2014neural}
Dzmitry Bahdanau, Kyunghyun Cho, and Yoshua Bengio.
\newblock Neural machine translation by jointly learning to align and
  translate.
\newblock In {\em 3rd International Conference on Learning Representations,
  {ICLR} 2015}, 2015.

\bibitem{liu2019point2sequence}
Xinhai Liu, Zhizhong Han, Yu-Shen Liu, and Matthias Zwicker.
\newblock Point2sequence: Learning the shape representation of 3d point clouds
  with an attention-based sequence to sequence network.
\newblock In {\em Proceedings of the AAAI Conference on Artificial
  Intelligence}, volume~33, pages 8778--8785, 2019.

\bibitem{li2018so}
Jiaxin Li, Ben~M Chen, and Gim Hee~Lee.
\newblock So-net: Self-organizing network for point cloud analysis.
\newblock In {\em Proceedings of the IEEE conference on computer vision and
  pattern recognition}, pages 9397--9406, 2018.

\bibitem{vaswani2017attention}
Ashish Vaswani, Noam Shazeer, Niki Parmar, Jakob Uszkoreit, Llion Jones,
  Aidan~N Gomez, {\L}ukasz Kaiser, and Illia Polosukhin.
\newblock Attention is all you need.
\newblock In {\em Advances in neural information processing systems}, pages
  5998--6008, 2017.

\bibitem{lee2019set}
Juho Lee, Yoonho Lee, Jungtaek Kim, Adam Kosiorek, Seungjin Choi, and Yee~Whye
  Teh.
\newblock Set transformer: A framework for attention-based
  permutation-invariant neural networks.
\newblock In {\em International Conference on Machine Learning}, pages
  3744--3753, 2019.

\bibitem{zhou2018voxelnet}
Yin Zhou and Oncel Tuzel.
\newblock Voxelnet: End-to-end learning for point cloud based 3d object
  detection.
\newblock In {\em Proceedings of the IEEE Conference on Computer Vision and
  Pattern Recognition}, pages 4490--4499, 2018.

\bibitem{wang2015voting}
Dominic~Zeng Wang and Ingmar Posner.
\newblock Voting for voting in online point cloud object detection.
\newblock In {\em Robotics: Science and Systems}, volume~1, pages 10--15607,
  2015.

\bibitem{li2016fpnn}
Yangyan Li, Soeren Pirk, Hao Su, Charles~R Qi, and Leonidas~J Guibas.
\newblock Fpnn: Field probing neural networks for 3d data.
\newblock In {\em Advances in Neural Information Processing Systems}, pages
  307--315, 2016.

\bibitem{riegler2017octnet}
Gernot Riegler, Ali Osman~Ulusoy, and Andreas Geiger.
\newblock Octnet: Learning deep 3d representations at high resolutions.
\newblock In {\em Proceedings of the IEEE Conference on Computer Vision and
  Pattern Recognition}, pages 3577--3586, 2017.

\bibitem{shi2015deeppano}
Baoguang Shi, Song Bai, Zhichao Zhou, and Xiang Bai.
\newblock Deeppano: Deep panoramic representation for 3-d shape recognition.
\newblock {\em IEEE Signal Processing Letters}, 22(12):2339--2343, 2015.

\bibitem{kanezaki2018rotationnet}
Asako Kanezaki, Yasuyuki Matsushita, and Yoshifumi Nishida.
\newblock Rotationnet: Joint object categorization and pose estimation using
  multiviews from unsupervised viewpoints.
\newblock In {\em Proceedings of the IEEE Conference on Computer Vision and
  Pattern Recognition}, pages 5010--5019, 2018.

\bibitem{su2018deeper}
Jong-Chyi Su, Matheus Gadelha, Rui Wang, and Subhransu Maji.
\newblock A deeper look at 3d shape classifiers.
\newblock In {\em Proceedings of the European Conference on Computer Vision
  (ECCV)}, pages 0--0, 2018.

\bibitem{qi2018frustum}
Charles~R Qi, Wei Liu, Chenxia Wu, Hao Su, and Leonidas~J Guibas.
\newblock Frustum pointnets for 3d object detection from rgb-d data.
\newblock In {\em Proceedings of the IEEE Conference on Computer Vision and
  Pattern Recognition}, pages 918--927, 2018.

\bibitem{thomas2019kpconv}
Hugues Thomas, Charles~R Qi, Jean-Emmanuel Deschaud, Beatriz Marcotegui,
  Fran{\c{c}}ois Goulette, and Leonidas~J Guibas.
\newblock Kpconv: Flexible and deformable convolution for point clouds.
\newblock In {\em Proceedings of the IEEE International Conference on Computer
  Vision}, pages 6411--6420, 2019.

\bibitem{xu2018spidercnn}
Yifan Xu, Tianqi Fan, Mingye Xu, Long Zeng, and Yu~Qiao.
\newblock Spidercnn: Deep learning on point sets with parameterized
  convolutional filters.
\newblock In {\em Proceedings of the European Conference on Computer Vision
  (ECCV)}, pages 87--102, 2018.

\bibitem{li2018pointcnn}
Yangyan Li, Rui Bu, Mingchao Sun, Wei Wu, Xinhan Di, and Baoquan Chen.
\newblock Pointcnn: Convolution on x-transformed points.
\newblock In {\em Advances in neural information processing systems}, pages
  820--830, 2018.

\bibitem{guo2020deep}
Yulan Guo, Hanyun Wang, Qingyong Hu, Hao Liu, Li~Liu, and Mohammed Bennamoun.
\newblock Deep learning for 3d point clouds: A survey.
\newblock {\em IEEE Transactions on Pattern Analysis and Machine Intelligence},
  2020.

\bibitem{luong2015effective}
Minh-Thang Luong, Hieu Pham, and Christopher~D Manning.
\newblock Effective approaches to attention-based neural machine translation.
\newblock {\em arXiv preprint arXiv:1508.04025}, 2015.

\bibitem{vinyals2015order}
Oriol Vinyals, Samy Bengio, and Manjunath Kudlur.
\newblock Order matters: Sequence to sequence for sets.
\newblock In {\em 4th International Conference on Learning Representations,
  {ICLR} 2016}, 2016.

\bibitem{xie2018attentional}
Saining Xie, Sainan Liu, Zeyu Chen, and Zhuowen Tu.
\newblock Attentional shapecontextnet for point cloud recognition.
\newblock In {\em Proceedings of the IEEE Conference on Computer Vision and
  Pattern Recognition}, pages 4606--4615, 2018.

\bibitem{yang2019modeling}
Jiancheng Yang, Qiang Zhang, Bingbing Ni, Linguo Li, Jinxian Liu, Mengdie Zhou,
  and Qi~Tian.
\newblock Modeling point clouds with self-attention and gumbel subset sampling.
\newblock In {\em Proceedings of the IEEE Conference on Computer Vision and
  Pattern Recognition}, pages 3323--3332, 2019.

\bibitem{9427563}
Zhiyong Tao, Yixin Zhu, Tong Wei, and Sen Lin.
\newblock Multi-head attentional point cloud classification and segmentation
  using strictly rotation-invariant representations.
\newblock {\em IEEE Access}, 9:71133--71144, 2021.

\bibitem{ba2016layer}
Jimmy~Lei Ba, Jamie~Ryan Kiros, and Geoffrey~E Hinton.
\newblock Layer normalization.
\newblock {\em arXiv preprint arXiv:1607.06450}, 2016.

\bibitem{wang2019dynamic}
Yue Wang, Yongbin Sun, Ziwei Liu, Sanjay~E Sarma, Michael~M Bronstein, and
  Justin~M Solomon.
\newblock Dynamic graph cnn for learning on point clouds.
\newblock {\em Acm Transactions On Graphics (tog)}, 38(5):1--12, 2019.

\bibitem{paszke2017automatic}
Adam Paszke, Sam Gross, Soumith Chintala, Gregory Chanan, Edward Yang, Zachary
  DeVito, Zeming Lin, Alban Desmaison, Luca Antiga, and Adam Lerer.
\newblock Automatic differentiation in pytorch.
\newblock 2017.

\bibitem{liu2019variance}
Liyuan Liu, Haoming Jiang, Pengcheng He, Weizhu Chen, Xiaodong Liu, Jianfeng
  Gao, and Jiawei Han.
\newblock On the variance of the adaptive learning rate and beyond.
\newblock {\em arXiv preprint arXiv:1908.03265}, 2019.

\bibitem{he2015delving}
Kaiming He, Xiangyu Zhang, Shaoqing Ren, and Jian Sun.
\newblock Delving deep into rectifiers: Surpassing human-level performance on
  imagenet classification.
\newblock In {\em Proceedings of the IEEE international conference on computer
  vision}, pages 1026--1034, 2015.

\bibitem{shapenet2015}
Angel~X. Chang, Thomas Funkhouser, Leonidas Guibas, Pat Hanrahan, Qixing Huang,
  Zimo Li, Silvio Savarese, Manolis Savva, Shuran Song, Hao Su, Jianxiong Xiao,
  Li~Yi, and Fisher Yu.
\newblock {ShapeNet: An Information-Rich 3D Model Repository}.
\newblock Technical Report arXiv:1512.03012 [cs.GR], Stanford University ---
  Princeton University --- Toyota Technological Institute at Chicago, 2015.

\bibitem{wang2020linformer}
Sinong Wang, Belinda Li, Madian Khabsa, Han Fang, and Hao Ma.
\newblock Linformer: Self-attention with linear complexity.
\newblock {\em arXiv preprint arXiv:2006.04768}, 2020.

\bibitem{xiong2021nystr}
Yunyang Xiong, Zhanpeng Zeng, Rudrasis Chakraborty, Mingxing Tan, Glenn Fung,
  Yin Li, and Vikas Singh.
\newblock Nystr\"omformer: A nystr\"om-based algorithm for approximating
  self-attention.
\newblock {\em arXiv preprint arXiv:2102.03902}, 2021.

\end{thebibliography}
\end{document}